\documentclass[runningheads]{llncs}
\usepackage{graphicx}

\usepackage{times}
\usepackage{latexsym}
\usepackage[utf8]{inputenc}
\usepackage{amssymb,amsfonts,amsmath}
\usepackage{graphicx} 
\usepackage{tabularx} 
\usepackage{multirow}
\usepackage{lipsum}
\usepackage{xcolor}
\usepackage{pgfplots,pgfplotstable}
\usepackage{pgfkeys}
\usepackage{float}
\usepackage{wrapfig}
\restylefloat{figure}
\usepackage{booktabs}
\usepackage[english]{babel}
\usepackage{caption}
\usepackage[flushleft]{threeparttable}

\usepackage{subcaption}
\usepackage{scrextend}
\let\oldFootnote\footnote
\newcommand\nextToken\relax

\renewcommand\footnote[1]{%
	\oldFootnote{#1}\futurelet\nextToken\isFootnote}

\newcommand\isFootnote{%
	\ifx\footnote\nextToken\textsuperscript{,}\fi}

\usepackage[linesnumbered,ruled]{algorithm2e}
\usepackage{setspace}
\usepackage{array}
\usepackage{calc}
\usepackage{url}

\newcommand{\x}{\mathbf{x}}
\newcommand{\y}{\mathbf{y}}

\newcommand{\p}{\mathrm{p}}

\newcommand{\VTheta}{\mathbf{\Theta}}

\usepackage[nameinlink,capitalize]{cleveref}

\DeclareMathOperator*{\argmax}{arg\,max~}

\definecolor{darkgreen}{RGB}{0,100,0}
\newcommand*{\regularbox}[1]{\color{black}
	\setlength{\fboxsep}{-1\fboxrule}
	\fbox{\hspace{1.1pt}\textbf{\strut#1}\hspace{1.2pt}}
	\color{black}  }

\begin{document}
\title{Interactive-predictive neural multimodal systems\thanks{The research leading to these results has received funding from MINECO under grant TIN2015-70924-C2-1-R, IDIFEDER/2018/025 "Sistemas de fabricación inteligentes para la industria 4.0" Actuación cofinanciada por la Unión Europea a través del Programa Operativo del Fondo Europeo de Desarrollo Regional (FEDER) de la Comunitat Valenciana 2014-2020, and from the European Commission under grant H2020, reference 825111. We also acknowledge NVIDIA Corporation for the donation of GPUs used in this work.}}

\author{Álvaro Peris \and Francisco Casacuberta}
\authorrunning{Á. Peris and F. Casacuberta}

\institute{Pattern Recognition and Human Language Technology Research Center \\ 
		   Universitat Politècnica de València, València, Spain \\
	 \email{\{lvapeab, fcn\}@prhlt.upv.es}
	 }

\maketitle

\begin{abstract}

Despite the advances achieved by neural models in sequence to sequence learning, exploited in a variety of tasks, they still make errors. In many use cases, these are corrected by a human expert in a posterior revision process. The interactive-predictive framework aims to minimize the human effort spent on this process by considering partial corrections for iteratively refining the hypothesis. In this work, we generalize the interactive-predictive approach, typically applied in to machine translation field, to tackle other multimodal problems namely, image and video captioning. We study the application of this framework to multimodal neural sequence to sequence models. We show that, following this framework, we approximately halve the effort spent for correcting the outputs generated by the automatic systems. Moreover, we deploy our systems in a publicly accessible demonstration, that allows to better understand the behavior of the interactive-predictive framework.

\end{abstract}

\section{Introduction}

The automatic prediction of structured objects is an extensively studied topic within the pattern recognition field. Many tasks involve the generation of a structured output, given an input object. As structure we understand a dependency across the elements of the object. Typical structured objects include sequences, trees or graphs. The application of neural networks to these problems has recently brought impressive advances. If both input and output objects are sequences, this problem is referred as sequence to sequence learning \cite{Graves12}. Many tasks can be posed as a sequence to sequence problem: machine translation \cite{Sutskever14}, speech recognition \cite{Chan16} or the automatic description of visual content, known as captioning \cite{Xu15,Yao15}. 

Notwithstanding the important breakthroughs achieved in the last years, these automatic systems are far from being error-free \cite{Koehn17}. However, they are useful for providing initial predictions, which are revised and corrected by a human expert. In some industries, such as machine translation, this revision procedure is widely used, as it increases the productivity with respect to performing the task from scratch \cite{Hu16}. This process is known as translation post-editing.

Nevertheless, this correction process can be improved in several ways. Aiming to increase the productivity of the system and seeking for a symbiotic human--computer collaboration, the so-called interactive-predictive pattern recognition was developed \cite{Foster97,Barrachina09}. Under this paradigm, the user introduces a correction to the system prediction. Next, the system reacts to this feedback, offering a new prediction, expected to be better than the previous one, as the system has more information. 

This interactive-predictive paradigm, initially devised for machine translation, can be extended to several tasks and technologies. In this work, we explore the application of this framework to several scenarios, which include data source from multiple modalities. In a nutshell, our main contributions are:

\begin{itemize}
	\item We successfully apply the interactive-predictive protocol to the automatic captioning of image and videos and to the machine translation post-editing, using neural sequence to sequence models. To the best of our knowledge, this is the first work that delves into this topic.
	\item We conduct experiments on several datasets, using two common neural architectures: a recurrent neural network (RNN) with attention and a Transformer model.
	\item We deploy our system in a freely accessible demonstration website.
	\item We release all the code developed in this work, fostering the research on this topic.	
\end{itemize}

The rest of the manuscript is structured as follows: in \cref{sec:ipr} we introduce the neural sequence to sequence modeling. Moreover, we describe the interactive-predictive pattern recognition framework and its implementation with neural models. Next, \cref{sec:experiments} details the experimental setup followed for assessing our systems. The evaluation and discussion of such systems are shown in \cref{sec:results}. \cref{sec:related-work} reviews the related work. Finally, in \cref{sec:conclusions} we extract conclusions and set the basis of future works.
\section{Interactive-predictive multimodal pattern recognition}
\label{sec:ipr}

The pattern recognition discipline consists in automatically obtaining a prediction $\hat{\y}$, given an input object $\x$.  A common approach to pattern recognition is based its statistical formalization. Following this probabilistic framework, the goal is to obtain the most likely prediction, given the input object:

\begin{equation}
\label{eq:pr}
\hat{\y} = \argmax_{\y} \Pr(\y \mid \x)
\end{equation}

Since the true probability distribution is unknown, it is approximated by a model with parameters $\VTheta$. Therefore, the prediction is given according to this model:

\begin{equation}
\label{eq:smt-eq}
\hat{\y} \approx \argmax_{\y} \p(\y \mid \x ; \VTheta)
\end{equation}

As aforementioned in the previous section, we are interested in the case in which both $\x$ and $\y$ are sequences. In the last years, and framed into the resurgence of neural networks, $\VTheta$ has been frequently implemented as a (deep) neural network, yielding the so-called neural sequence to sequence modeling. This neural network is usually trained on an end-to-end manner on large datasets, via stochastic gradient descent. Moreover, since performing a complete search is prohibitively expensive, the $\argmax$ is solved by applying a heuristic search method, typically, beam search \cite{Sutskever14}.

\subsection{Neural architectures for multimodal sequence to sequence learning}

\begin{figure*}[!h]
	\centering
	\begin{subfigure}[b]{0.35\textwidth}
		\includegraphics[width=\textwidth]{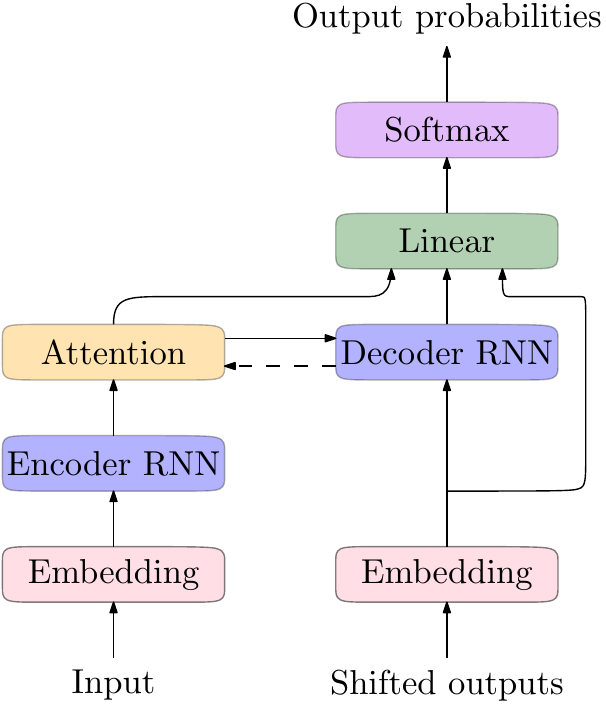}
		\caption{RNN-based encoder--decoder with attention.}
		\label{fig:rnn-system}
	\end{subfigure}
	\qquad\qquad  
	\begin{subfigure}[b]{0.44\textwidth}
		\includegraphics[width=\textwidth]{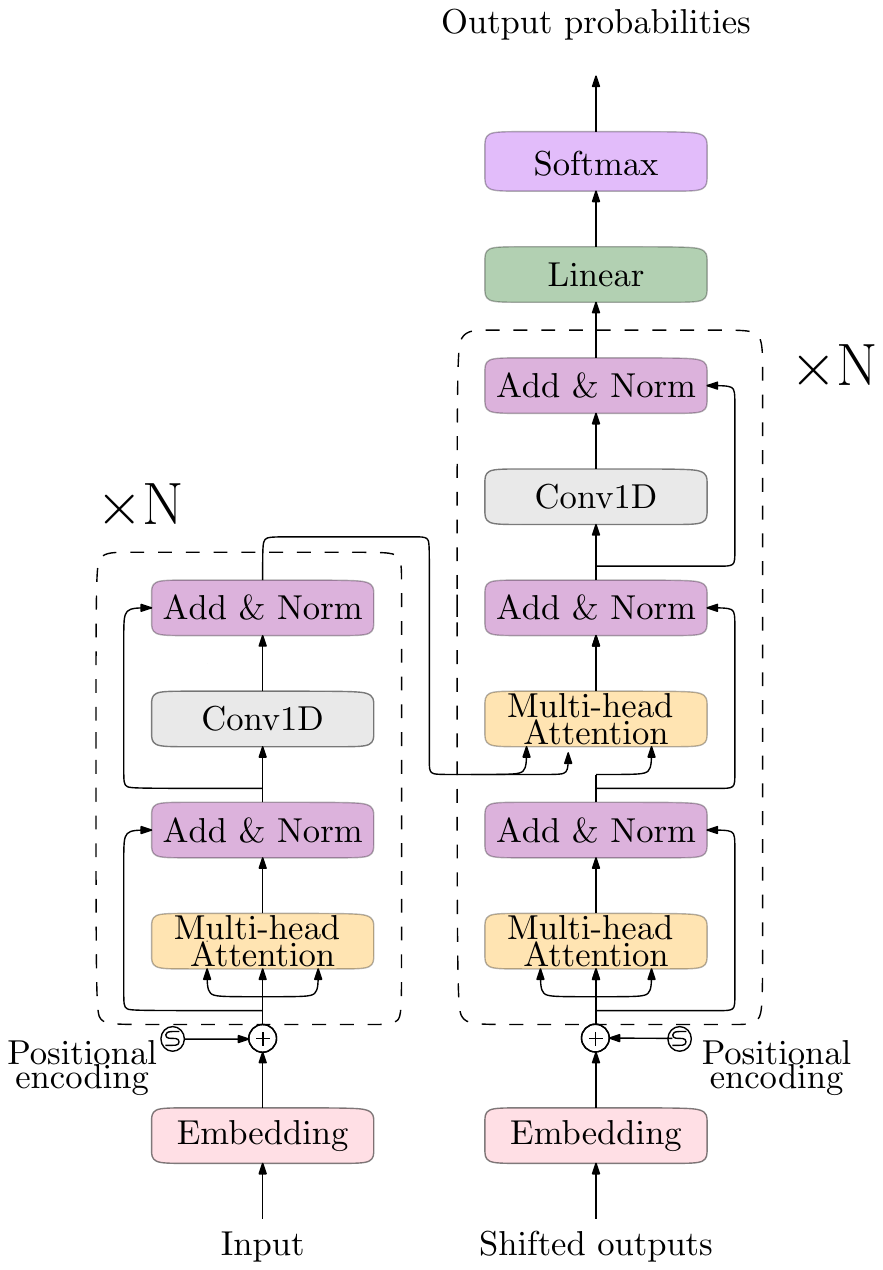}
		\caption{Transformer model, as illustrated by \cite{Vaswani17}.}
		\label{fig:transformer-system}
	\end{subfigure}
	\caption{\label{fig:encoder-decoder}Different architectures for sequence to sequence learning: RNN-based (left) and Transformer models (right). Both models have the same inputs and outputs and differ on the mechanisms applied for learning their representations. In the first case, the input sequence is analyzed by an encoder RNN. The output sequence is generated, word by word, by another RNN. Both RNNs are connected through an attention mechanism. In the case of the Transformer model, the encoder and the decoder are stacks of multi-head attention mechanisms that compute different representations of the inputs. Both models have a vocabulary-sized output layer with a softmax activation, that computes a probability distribution over the output vocabulary.}
\end{figure*}

Most neural models for sequence to sequence learning rely on the encoder--decoder paradigm: first, a neural encoder computes a representation of the input sequence. Next, a neural decoder takes this representation is then generates, element by element, the output sequence. Alternative architectures for encoder and decoder have been proposed in the literature. The most popular among them are those based on RNNs with attention \cite{Bahdanau15} or those based solely on attention mechanisms \cite{Vaswani17} (the so-called Transformer models). \cref{fig:encoder-decoder} depicts a schematic view of these systems. However, providing an in-depth review of these models is out of the scope of this paper. Hence, we refer the reader to the original works for a detailed explanation of these architectures.

This encoder--decoder paradigm can be applied to sequences from arbitrary sources. The only requirement is that we need to encode the input object into a low dimensional, real-valued representation. In this work, we focus on objects from three different sources: text, images and video. Hence, before being introduced to the encoder--decoder system, we need to compute an adequate representation of them. In the computer vision field, this process is known as feature extraction.  Depending on the modality of the input object, we thus apply a different feature extractor:

\begin{description}
	\item [Text:] each word is mapped to a continuous representation by using an embedding matrix \cite{Sutskever14}. Hence, the sequence of input words is converted to a sequence of word embeddings. The embedding matrix is usually estimated with the rest of the parameters of the model.
	
	\item [Images:] convolutional neural networks (ConvNets, \cite{LeCun98}) excel in several computer vision tasks \cite{Krizhevsky12}. These models are also powerful feature extractors. We process the image with a ConvNet and use as features the final representation computed by the ConvNet that preserves positional information. A complete image is thus seen as a sequence of image crops. Hence, we can directly apply the sequence to sequence framework, as done by Xu et al. \cite{Xu15}.
	
	\item [Videos:]	A video is a sequence of images. Therefore, we also rely on the usage of ConvNet for extracting the features from the each video frame. For alleviating the computational overload, we compute global features for each video image. In addition, we subsample the frames introduced to the system \cite{Yao15}, also for reducing the computational load.
 \end{description}

\subsection{Interactive-predictive pattern recognition}

As discussed in the previous section, in an interactive-predictive scenario, the user introduces corrections to the predictions generated by a pattern recognition system. This correction is introduced as a feedback signal $f$. The systems reacts then to the introduction of the feedback, producing an alternative hypothesis, compatible with $f$. Considering this, the interactive-predictive framework rewrites \cref{eq:smt-eq} for also taking into account the user feedback signal:

\begin{equation}
\label{eq:imt}
\tilde{\y}  = \argmax_{\y~\textrm{compatible with}~f} \p(\y \mid \x, f; \VTheta)
\end{equation}

Hence, the goal of an interactive-predictive system is to generate the most likely prediction that is compatible with the feedback provided by the user. Depending on the meaning conveyed by $f$, alternative interactive protocols can be defined. In this work, we follow the prefix-based interactive protocol. We also assume that the user introduces the corrections using a keyboard and a mouse. 

The prefix-based protocol arguably is the most natural way of work. In this protocol, the user searches, from the left to the right, for the first error in the prediction given by the system and introduces the correct character. This feedback signal conveys a two-fold meaning: on the one hand, it states a correct character at a given position. On the other hand, it also validates the hypothesis up to this position. Taking this into account, a prefix-based interactive-predictive system must generate the most likely suffix, to a prefix validated by the user \cite{Barrachina09}.

The implementation of this protocol in neural sequence to sequence systems requires to constrain the search \cite{Peris17a}: the system applies a forced decoding of the feedback provided by the user. The suffix is obtained then by applying a regular search. For introducing corrections at a character level, we apply a vocabulary mask as described by \cite{Peris19}, which ensures that the next word generated complies with the user feedback.

\label{sec:experiments}

We evaluate our interactive-predictive framework in six different scenarios, involving three tasks and two different datasets per task. The main figures of the datasets are shown in \cref{table:datasets}. The tasks under study are:

\begin{description}
	\item [Machine translation:]translation of English sentences to French, on two datasets\footnote{Datasets available at: \url{http://statmt.org/wmt18}}: UFAL and Europarl. The first one belongs to a medical domain and the latter refers to the translation of the proceedings from the European parliament.

	\item [Image captioning:]we tackled two common datasets: Flickr8k \cite{Hodosh13} and Flickr30k \cite{Plummer15}. The goal is to generate descriptions of pictures crawled from Flickr users. 
	
	\item [Video captioning:]we tested our systems on the popular Microsoft Research Video Description (MSVD) dataset \cite{Chen11}, a general task, relating the description of YouTube videos from multiple domains. In addition, we apply our methods to the EDUB-SegDesc dataset \cite{Bolanos18}, a collection of egocentric videos and first person captions.
\end{description}

\begin{table}[!h]
	\caption {\label{table:datasets}Figures of the different datasets. $M$ denotes millions of elements. The column \#References indicates the number of different references per sample. $^\star$ denotes a variable number of references. In this case, we report the average references per sample. }
	\centering

	\begin{tabular*}{\textwidth}{l@{\extracolsep{\fill}}lrrrc}
		\toprule
		\multirow{2}{*}{Task} & \multirow{2}{*}{Dataset} & \multicolumn{3}{c}{\#Samples} & \multirow{2}{*}{\#References} \\
		\cmidrule(lr){3-5} 
		& & Training & Validation & Test & \\
		\midrule
		\multicolumn{1}{ l }{\multirow{2}{*}{Machine Translation}} 
		& UFAL & $2.8M$ & $1,000$ & $1,000$ & $1$\\ 
		& Europarl & $2.0M$ & $3,003$ & $3,000$& $1$ \\ 
		\midrule
		\multicolumn{1}{ l }{\multirow{2}{*}{Image Captioning}} 
		& Flickr8k & $30,000$ & $1,000$ & $1,000$ & $5$ \\ 
		& Flickr30k & $145,000$ & $1,014$ & $1,000$& $5$ \\ 
		
		\midrule
		\multicolumn{1}{ l }{\multirow{2}{*}{Video Captioning}} 
		& MSVD & $48,779$ & $100$ & $670$& $41^\star$\\ 
		& EDUB-SegDesc & $2,652$ & $204$ & $246$ & $3$\\ 
		\bottomrule
	\end{tabular*}
\end{table}

\subsection{Evaluation metrics}

We evaluate two main aspects of our systems. On the one hand, we measure the quality of the initial predictions provided by the system. This is the most common scenario in the literature. This evaluation is carried on by comparing the predictions with the ground-truth references from each dataset. The final goal of these metrics is to correlate with the human perception of prediction quality. The metrics range from $0$ (worst quality) to $100$ (best quality):

\begin{description}
	\item [BLEU \cite{Papineni02}:]Computes the geometric mean of the $n$-gram precision of prediction and references. In includes $n$-grams from order $1$ to $4$. It also includes a penalty for short predictions.
	\item [METEOR \cite{Lavie09}:]Computes the F1 score of precision and recall of matches between prediction and references words. To this end, it applies linguistic resources such as stemmers, paraphrase and synonym dictionaries.
\end{description}

On the other hand, under an interactive-predictive framework, our objective is to reduced the amount of effort spent by the user during the correction process. We follow the literature and estimate this effort as the number of keystrokes and mouse actions performed by the user during the correction process. To this end, we rely on two metrics:

\begin{description}
	\item [CharacTER \cite{Wang16b}:]Translation edit rate computed at a character level: minimum number of character edit operations (insertion, substitution, deletion and swapping) that must be made in order to transform the hypothesis into the reference. The number of edit operations is normalized by the number of characters.
	\item [KSMR \cite{Barrachina09}:]accounts for the number of keystrokes plus mouse actions involved in the interactive correction process, divided by the number of characters of the final prediction obtained.
\end{description}

CharacTER and KSMR are error-based metrics, hence the lower, the better. Following Zaidan et al. \cite{Zaidan10}, CharacTER is an estimate of the effort of static post-edition; while the effort of interactive-predictive systems can be assessed via KSMR \cite{Barrachina09}.

\subsection{Usage of the system and user simulation}

Using an interactive-predictive system requires to follow the procedure described in \cref{sec:ipr}: the process starts with an automatic prediction given by the system to an input object. The user then reviews the prediction, starting and the interactive-predictive process: the user searches in this hypothesis the first error, and introduces a correction. The system then reacts, providing an alternative hypothesis, considering the user feedback. 
This protocol is repeated until the user finds satisfactory the hypothesis given by the system. We implemented a live demonstration of this system\footnote{Accessible at \url{http://casmacat.prhlt.upv.es/interactive-seq2seq}}. 

Properly assessing interactive-predictive systems involves the experimentation with human users, which is prohibitively expensive. Hence, during the development of such systems, it is common to rely on simulated users \cite{Barrachina09,Peris17a}. We used the ground-truth samples from the different datasets as the desired outputs by our simulated users. The simulation is done by correcting the leftmost wrong character of each hypothesis from the interactive-predictive system, until reaching the desired output.

\subsection{Description of the systems}

Our neural sequence to sequence systems\footnote{Source code: \url{https://github.com/lvapeab/interactive-keras-captioning}} were developed with NMT-Keras \cite{Peris18b}. This library is built upon Keras\footnote{\url{https://keras.io}} and works for the Theano and Tensorflow backends. For each task and dataset, we built two models: one using RNNs with attention and another one using a Transformer architecture.

The RNN-based systems had long short-term memory units \cite{Hochreiter97}. Encoder and decoder were bridged together through an additive attention mechanism \cite{Bahdanau15}. We set all model dimensions to the same value. In the case of machine translation, all layers had a dimension of $512$. In the case of image and video captioning, we reduced the model size to $256$, since we are dealing with smaller datasets.

In the case of the Transformer models, we set two stacks of $6$ layers for the encoder and the decoder. In the case of machine translation, all model dimensions were $512$ and the number of attention heads was $8$. This configuration is the same as the \textit{base} model described by Vaswani et al. \cite{Vaswani17}. For the captioning tasks, we reduced again our model, to $256$ dimensions on each layer.

Machine translation and image captioning systems were trained using Adam \cite{Kingma14}, with a learning rate of $0.0002$. In the case of video captioning, we obtained better performance using Adadelta \cite{Zeiler12}, in both datasets. In all cases, the batch size was $64$. During training, we applied an early-stopping strategy, watching the BLEU on the development set. At decoding time, we used a beam size of $6$.

In the case of machine translation, the word embeddings were randomly initialized and learned together with the rest of the parameters of the system. In the case of image captioning, we extracted image features using a NASNet architecture \cite{Zoph18}, trained on the ImageNet dataset \cite{Deng09}. The video features were extracted with an Inception v4 network \cite{Szegedy16}, also trained on the ImageNet dataset. Following Yao et al. \cite{Yao15}, we subsampled the frames from a video, selecting 26 images per clip. Image and video feature remained static along the training process of the sequence to sequence model.
\section{Results and discussion}
\label{sec:results}

We show and discuss now the results obtained by our systems. First, we will assess the systems quantitatively, in terms of prediction quality and effort required during the correction stage. Next, in order to gain some insights into the behavior of the system, we analyze an image captioning example.

\subsection{Quantitative evaluation}

We start by evaluating the systems in a traditional way, assessing their prediction quality. \cref{table:results-mt} shows the BLEU and METEOR results of the different systems for all tasks. These results are similar to those reported in the literature for each task and dataset \cite{Bolanos18,Peris19,Xu15,Yao15}.
\begin{table}[h]
	\caption {\label{table:results-mt}Prediction quality for the different tasks, datasets and models. The RNN column denotes RNN-based system (\cref{fig:rnn-system}) and the Trans. column indicates a Transformer model (\cref{fig:transformer-system}).}
	\centering

	\begin{tabular*}{\textwidth}{l@{\extracolsep{\fill}}lcccc}
		\toprule
		\multirow{2}{*}{Task} & \multirow{2}{*}{Dataset} & \multicolumn{2}{c}{BLEU~[$\uparrow$]} 
		& \multicolumn{2}{c}{ METEOR~[$\uparrow$]} 
		\\ \cmidrule(lr){3-4}\cmidrule(lr){5-6} 
		& & \multicolumn{1}{c}{RNN} & \multicolumn{1}{c}{Trans.}  & \multicolumn{1}{c}{RNN} & \multicolumn{1}{c}{Trans.}  \\
		\midrule
		\multicolumn{1}{ l }{\multirow{2}{*}{Machine Translation}} 
	      & UFAL & $37.2$ & $37.8$&  $59.6$ & $60.4$ \\ 
    	  & Europarl & $24.6$ & $26.6$ &  $45.7$ & $47.9$ \\ 
		\midrule
		\multicolumn{1}{ l }{\multirow{2}{*}{Image Captioning}} 
		& Flickr8k & $22.1$ & $19.6$&  $20.8$ & $19.8$ \\ 
		& Flickr30k & $22.2$ & $19.3$ &  $20.0$ & $18.5$ \\ 

		\midrule
		\multicolumn{1}{ l }{\multirow{2}{*}{Video Captioning}} 
		& MSVD & $49.6$ & $45.7$&  $33.4$ & $30.7$ \\ 
		& EDUB-SegDesc & $30.4$ & $25.8$ &  $21.9$ & $20.3$ \\ 
		\bottomrule
	\end{tabular*}
\end{table}

It is worth to note that the Transformer model only outperformed the RNN-based systems in the case of machine translation. This model is more data-eager than RNN systems. Many of the recent advances yielded with this architecture leverage huge data collections (e.g. Radford et al. \cite{Radford19}). We also contrasted this fact in our experimentation: the machine translation datasets were way larger than the captioning ones (see \cref{table:datasets}. Hence, the Transformer model only was fully exploited in the machine translation case.


Next, we evaluate the performance of the interactive-predictive systems. To that end, we estimate the effort required for correcting the output of a static system (using CharacTER) and the effort needed by a interactive system (using KSMR). These results are shown in \cref{table:results-ksmr}. The results obtained in machine translation are similar to the literature \cite{Peris19}. Due to the novelty of this scenario, we lack from references in the literature, regarding the other tasks.

\begin{table}
	\caption[KSMR of interactive systems at a character level]{\label{table:results-ksmr}Effort required for correcting the outputs of static (St.) and interactive-predictive (Int.) systems, using RNN and Transformer (Trans.) models. The effort of static systems is measured in terms of CharacTER while the effort required by interactive-predictive systems is evaluated in terms of KSMR.} 
	\centering

	\begin{tabular*}{\textwidth}{l@{\extracolsep{\fill}}lcccc}
		\toprule
		\multirow{2}{*}{Task} & \multirow{2}{*}{Dataset} & \multicolumn{2}{c}{CharacTER~[$\downarrow$]} & \multicolumn{2}{c}{KSMR~[$\downarrow$]} 
		\\ \cmidrule(lr){3-4} \cmidrule(lr){5-6}
		& & \multicolumn{1}{c}{St. RNN} & \multicolumn{1}{c}{St. Trans.}    & \multicolumn{1}{c}{Int. RNN} & \multicolumn{1}{c}{Int. Trans.}   \\
		\midrule
		\multicolumn{1}{ l }{\multirow{2}{*}{Machine Translation}} 
		& UFAL & $35.7$ & $36.5$ & $19.0$ & $15.9$  \\ 
		& Europarl & $53.6$ & $51.2$ & $30.1$ & $29.4$  \\ 
		\midrule
		\multicolumn{1}{ l }{\multirow{2}{*}{Image Captioning}} 
		& Flickr8k & $77.8$ & $79.6$ &$36.6$ & $36.9$ \\ 
		& Flickr30k &$81.7$ & $86.1$& $36.0$ & $40.0$ \\ 
		
		\midrule
		\multicolumn{1}{ l }{\multirow{2}{*}{Video Captioning}} 
		& MSVD & $58.1$ & $64.1$ &$36.4$ & $40.5$ \\ 
		& EDUB-SegDesc  & $72.3$ & $71.4$ & $40.0$ & $38.0$ \\ 
		\bottomrule
	\end{tabular*}
\end{table}

Interactive-predictive systems approximately halved the amount of corrections required for correcting their outputs, with respect to traditional, static systems. The results were consistent across all tasks and for all models. Hence, these results indicate that the interactive protocol effectively achieved its goal of reducing the correction effort.

Moreover, a crucial aspect of the usability of interactive systems is their response time. Hence, it is important to keep it in adequate values. The average response time of our systems was always below $0.2$ seconds. This provides the user of a feeling of almost instant reactivity \cite{Nielsen93}.

Finally, we are aware that properly assessing the usability and effort reduction brought by these system requires a human evaluation on its usage. In this paper, we set the first step toward future developments on multimodal neural interactive-predictive pattern recognition, with positive initial results. 

\subsection{Qualitative analysis and discussion}

\begin{figure*}[!h]
	\centering
	\includegraphics[width=0.8\textwidth]{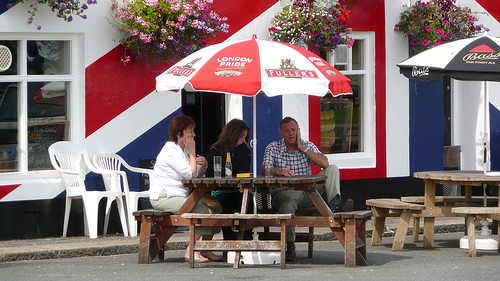}\vspace*{0.5cm}
	\small
	\begin{tabular*}{\textwidth}{l@{\extracolsep{\fill}}ll}
		\toprule
		\textbf{Iter 0} & \textbf{System} & A group of people sit on a ramp. \\
		\midrule
		\multirow{2}{*}{\textbf{Iter 1}} & \textbf{User} & \textcolor{darkgreen}{\textit{A group of people sit on a }} \regularbox{b}ramp. \\
		& \textbf{System} &\textcolor{darkgreen}{ \textit{A group of people sit on a b}}ench. 
		\\
		\midrule
		\multirow{2}{*}{\textbf{Iter 2}} & \textbf{User} & \textcolor{darkgreen}{\textit{A group of people sit on a bench}} \regularbox{u}. \\
		& \textbf{System} &\textcolor{darkgreen}{ \textit{A group of people sit on a bench u}}nder a building. \\
		\midrule
		\multirow{2}{*}{\textbf{Iter 3}} & \textbf{User} & \textcolor{darkgreen}{\textit{A group of people sit on a bench under a}}\regularbox{n}building. \\
		& \textbf{System} &\textcolor{darkgreen}{ \textit{A group of people sit on a bench under an}} umbrella. \\
		\midrule
		\multirow{1}{*}{\textbf{Iter 4}} & \textbf{User} & \textcolor{darkgreen}{\textit{A group of people sit on a bench under an umbrella.}} \\
		
		\bottomrule
	\end{tabular*}
	
	\caption{\label{fig:exampleIMT} Interactive-predictive session example, for correcting the caption generated for the image. At each iteration, the user introduces a character correction (boxed). The system modifies its hypothesis, taking into account this feedback: keeping the correct prefix (green) and generating a compatible suffix. Post-editing this sample in a static way, would have required the deletion of $4$ characters and the addition of $23$ characters. }
\end{figure*}

We show and analyze an image captioning example. Other examples for the machine translation and video captioning tasks are alike. The example is taken from our multimodal showcase and shown in \cref{fig:exampleIMT}.

We can see that the caption generated by the system (at iteration $0$) has an error. The user wants to indicate that the people are sitting on a bench. Hence, the feedback introduced is the character ``b''. The system is able to properly complete the word ``bench'', with this single interaction. The same happens when the user wants to introduce the clause ``under a''. With only typing the character ``u'', the system generates this clause. Finally, it is interesting to observe the behavior of the last interaction. The user introduced the character ``n'' to the word ``a''. Hence, the next word must start with a vowel. The system is able to properly account for this concordance and generates the word ``umbrella''. We observe that the systems also handle correctly other concordances, such as singular/plural clauses. 


\section{Related work}
\label{sec:related-work}

Neural sequence to sequence learning has been a widely studied topic since its reintroduction, framed to the deep learning era \cite{Graves12,Sutskever14}. As stated above, neural machine translation \cite{Bahdanau15,Vaswani17} has meant a revolution in the field. Nowadays, these systems are standard in research and industry. In addition to machine translation, different tasks have been tackled following this approach: speech recognition \cite{Chan16}, speech translation \cite{Jia19}, syntactic parsing \cite{Vinyals15b}, or the already discussed image and video captioning \cite{Xu15,Peris16,Yao15}. 

Regarding the interactive-predictive pattern recognition framework, it has been mainly applied to machine translation. The addition of interactive protocols for fostering the productivity of translation environments have been studied for long time, for phrase-based models \cite{Barrachina09,Green14} and neural machine translation systems \cite{Knowles16,Peris17a}. 

The interactive-predictive approach has been also previously generalized for tackling other tasks, involving multimodal signals. This is the case of the interactive transcription of handwritten text documents \cite{Toselli07}, layout detection \cite{Quiros17}, among others \cite{Toselli11}. None of these works however, involved fully end-to-end neural multimodal systems.

\section{Conclusions and future work}
\label{sec:conclusions}

In this work, we empirically demonstrated the capabilities of the interactive-predictive framework applied to multimodal, neural sequence-to-sequence systems. We tackled a variety of tasks, using two state-of-the-art models and, in all cases, the interactive-predictive systems were able to decrease the human effort required for correcting the outputs of the system. We obtained savings of approximately a $50\%$. We also analyzed these systems through an online demo website. We released all source code developed.

These encouraging results open several avenues for future research. The construction of multimodal, interactive-predictive systems allow the application of this framework to other structured prediction tasks, e.g. tables to text. More precisely, this framework is directly applicable to the automatic report of medical images or to the automatic generation of life-loggers. In addition to an end application, these tools can be used by human annotators, for creating datasets on a more efficient way.

Moreover, we experimented with multimodal inputs. In a future, we want to explore the inclusion of multimodal feedback signals. This was already done for statistical models \cite{Alabau14} and we think that neural models are able to exploit this very effectively. In addition, we used a different system for each task. In a future, we would like to explore the construction of a single multitask, multimodal system. The recent advances achieved in multitask learning \cite{Radford19} heavily support this research direction. Finally, for properly assessing the efficiency of this framework, we should conduct and experimentation involving human users.

\bibliographystyle{splncs04}
\bibliography{ibpria}

\end{document}